\documentclass{article}
\usepackage{ijcai26} 

\usepackage{times}
\usepackage{soul}
\usepackage{url}
\usepackage[hidelinks]{hyperref}
\usepackage[utf8]{inputenc}
\usepackage[small]{caption}
\usepackage{graphicx}
\usepackage{tabularx}
\usepackage{adjustbox}
\usepackage{amsmath}
\usepackage{amsthm}
\usepackage{booktabs}
\usepackage{algorithm}
\usepackage{algorithmic}
\usepackage[table]{xcolor}
\usepackage[most]{tcolorbox}
\usepackage[switch]{lineno}
\newcommand{\cmark}{\ding{51}} 
\newcommand{\xmark}{\ding{55}} 
\PassOptionsToPackage{table}{xcolor}
\usepackage{float}
\usepackage{pifont}
\usepackage{amsfonts}

\title{Less is More: Modality-Decoupling for General AIGC Audio-Video Detection}

\author{
Jielun Peng$^1$
\and
Yabin Wang$^1$
\and
Yaqi Li$^1$
\and
Jincheng Liu$^1$\and
Xiaopeng Hong$^{1}$\thanks{Corresponding author.}\\ \And
Athanasios V. Vasilakos$^{2}$\\
\affiliations
$^1$Harbin Institute of Technology \quad $^2$University of Agder
\emails
\{25s003052, 25s103223, 25b903114\}@stu.hit.edu.cn,
wang-yabin@outlook.com,
hongxiaopeng@ieee.org, th.vasilakos@gmail.com
}

\begin{document}
\maketitle

\begin{abstract}
Generative AI has rapidly expanded audio-visual forgery beyond human-centric deepfakes into general scenes.
Existing AIGC detection methods assume audio-visual content correspondence, identifying forgeries by spotting cross-modal inconsistencies. 
However, we empirically find that this assumption does not consistently hold in general scenarios.
We argue that, for general audio-visual AIGC detection, decision-level fusion is a more robust alternative to feature-level fusion.
Therefore, we propose DAV-Det, a decoupled audio-visual AIGC detection system that independently models forensic evidence from each modality.
The visual detector leverages multi-granularity representations at global, patch, and segment levels to capture spatial forgery cues, while the audio detector exploits both temporal and spectral irregularities via a gated temporal-spectral dual-branch architecture to model acoustic artifacts. 
Our method ranks 1st in the General AIGC Audio-Video Detection Challenge of the IJCAI-ECAI 2026 DDL 2.0 Workshop, with a final score of 0.8460.
Code is available at \textit{\url{https://github.com/tuffy-studio/DAV-Det}}.
\end{abstract}    
\section{Introduction}
\label{sec:intro}

The rapid advancement of Artificial Intelligence Generated Content (AIGC) technologies has revolutionized the creation of audio-visual content. While early AIGC systems primarily focused on human-centric generation, recent advances in generative models have enabled the synthesis of realistic audio-visual content in diverse scenarios beyond human faces and speech~\cite{wang2024linguistic,wang2025benchmark,demamba}. This shift substantially broadens the scope of potential misuse, ranging from misinformation to forged evidence. Consequently, reliable detection of general AIGC content has become increasingly important.

\begin{figure}[t]
    \centering
    \includegraphics[width=1\linewidth]{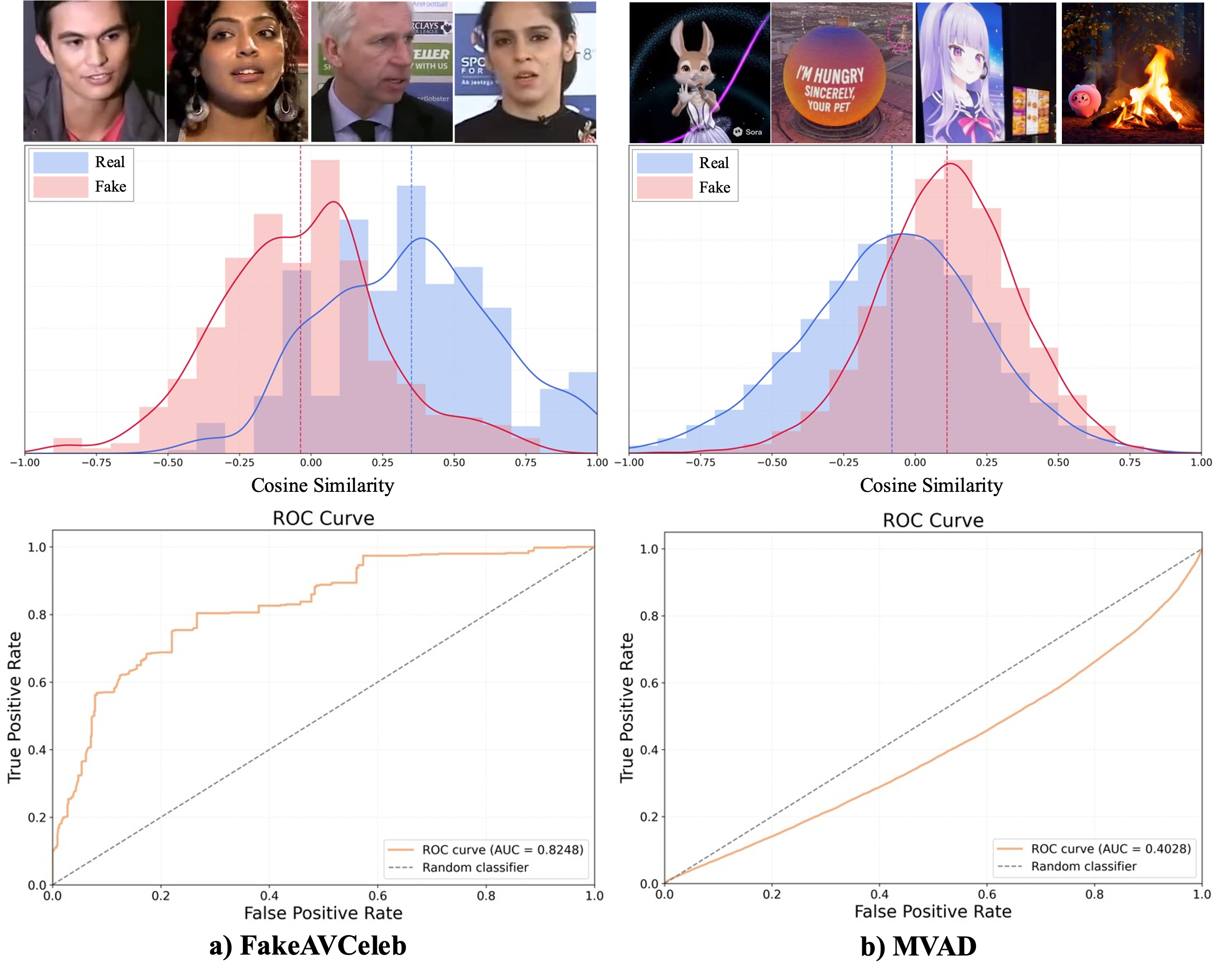}
    \vspace{-2em}
    \caption{Distributions of cosine similarities between audio and video features extracted by $\text{PEAV}$. On FakeAVCeleb, real samples exhibit higher similarity than fake samples. In contrast, the audio-visual similarities significantly decrease in MVAD compared to human-centric datasets. Moreover, the similarities of real samples are generally even lower than those of fake ones in MVAD. Using this score alone for training-free detection yields an AUC below 0.5, suggesting that the traditional assumption of audio-visual correspondence is actively detrimental in general AIGC scenarios.}
    \vspace{-1em}
    \label{fig:teaser}
\end{figure}
To combat the growing threat of AI-generated forgeries, numerous detection methods have been proposed. Some approaches focus on a single modality, detecting artifacts only from either audio~\cite{aasist,rawbmamba,allm4add} or visual modalities~\cite{opensdi,poundnet}. Such methods fail to fully exploit the comprehensive information available across modalities. 
Building upon uni-modal limitations, multi-modal detection approaches~\cite{AVgraph,avprompt,avff} have achieved remarkable success by jointly modeling audio and visual information. 
These methods are typically based on the assumption that, in human-centric scenarios, audio and visual signals exhibit content correspondence between speech phonemes and lip movements. Such correspondence is typically assessed via audio-visual matching measures, including frame-level alignment~\cite{avad,avh-align} and feature similarity~\cite{speechforensics,havic}. Manipulation of either modality may disrupt this correspondence, thereby providing discriminative cues for AIGC detection.

However, this assumption does not consistently hold in general audio-visual scenarios. While human-centric content often exhibits content consistency between audio and visual streams, general videos tend to show weaker alignment in their underlying content.
To validate our hypothesis, we use PEAV~\cite{PEAV} to quantify audio-visual correspondence on two representative datasets: FakeAVCeleb~\cite{fakeavceleb} and MVAD~\cite{mvad}. FakeAVCeleb is a widely used human-centric deepfake dataset, while MVAD contains general-scene AIGC content. 
As shown in Fig.~\ref{fig:teaser}, real samples in FakeAVCeleb exhibit higher audio-visual cosine similarity than fake samples. 
Following~\cite{speechforensics,avh-align}, we rank samples based on the negative of cosine similarity and compute the area under the receiver operating characteristic curve (AUC), achieving a score of 82.48\%. This result indicates that audio-visual correspondence serves as an effective cue for human-centric deepfake detection.
However, on MVAD, fake samples even show higher similarity than real ones. Under the same evaluation protocol, where fake samples are expected to have lower audio-visual similarity, this leads to an AUC of only 40.28\% (i.e., even worse than random guessing). 
As a result, the effectiveness of existing multi-modal detectors may be limited in general AIGC scenarios.

\begin{figure*}[t]
    \centering
    \includegraphics[width=1.0\linewidth]{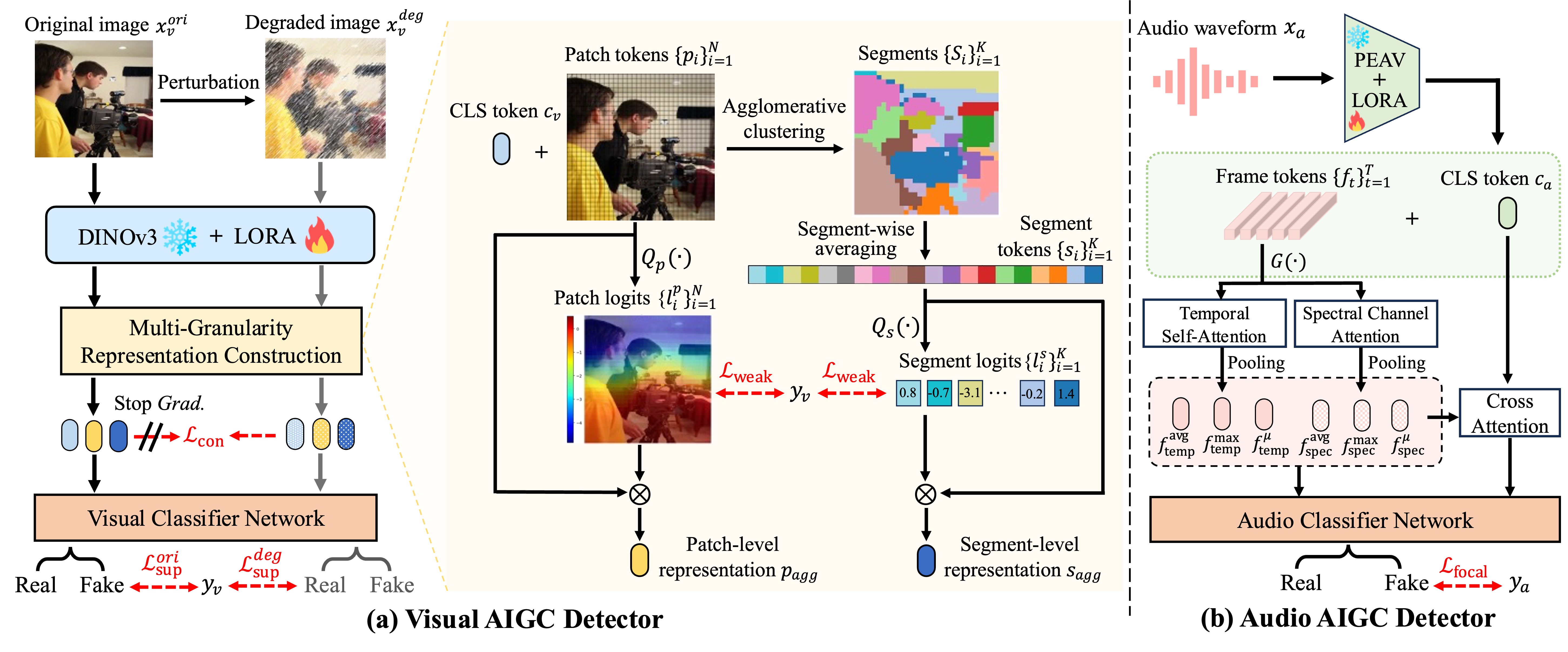}
    \vspace{-2em}
    \caption{\textbf{Overview of our proposed DAV-Det.} (a) The visual AIGC detector models global, patch-, and segment-level representations to capture forgery cues at multiple spatial granularities. (b) The audio AIGC detector characterizes temporal dynamics and spectral anomalies in the audio stream for comprehensive AIGC detection.}
    \label{fig:framework}
    \vspace{-1em}
\end{figure*}

These findings motivate us to explore a simpler yet more robust paradigm: independently modeling modality-specific forensic cues and performing decision-level fusion. Accordingly, we propose DAV-Det, a \underline{D}ecoupled \underline{A}udio-\underline{V}isual AIGC \underline{Det}ection system that avoids explicit feature-level cross-modal interaction. Specifically, considering that AIGC cues may emerge at multiple spatial granularities, including global semantic anomalies, subtle local texture artifacts and regional inconsistencies, we develop a visual AIGC detector (see Fig.~\ref{fig:framework} (a)) that jointly exploits global, patch-, and segment-level representations to capture forgery cues at different levels. Complementarily, we design an audio AIGC detector (see Fig.~\ref{fig:framework} (b)) that exploits both temporal dependencies and spectral irregularities via a gated temporal-spectral dual-branch architecture, enabling fine-grained modeling of acoustic artifacts. During inference, the two detectors operate independently and their predictions are fused to enable both binary and four-class classification.

Our main contributions are summarized as follows:
\begin{itemize}
    \item We empirically demonstrate that the assumption of audio-visual content correspondence, which underlies most existing multi-modal deepfake detectors, does not consistently hold in general AIGC scenarios.
    \item We propose DAV-Det, which independently models audio and visual forensic cues via multi-granularity visual representations and an audio temporal-spectral design, and performs decision-level fusion for audio-visual AIGC detection.
    \item Our method demonstrates competitive performance in the General AIGC Audio-Video Detection Challenge of the DDL 2.0 Workshop at IJCAI-ECAI 2026, ranking 1st with a final score of 0.8460.
\end{itemize}
\section{Related Works}
\label{sec:rw}
\subsection{Uni-modal AIGC Detection}
\noindent\textbf{Visual-only methods.}
Methods leveraging visual artifacts focus on spatial and temporal inconsistencies, such as abnormal regions~\cite{Multi_attentional_deepfake_detection} and irregular movements~\cite{lip-forensics}.
To improve generalization, prior works explore data augmentation with synthetic samples. Representative methods include artifact simulation, e.g., blending boundaries~\cite{SBI,sladd}, and latent-space augmentation to enhance decision boundary~\cite{LSDA,simlbr}.
Another line of work is leveraging the pre-trained models to alleviate overfitting. For example, Effort~\cite{effort} decomposes the feature space into orthogonal subspaces, preserving pre-trained knowledge while learning forgery cues. Query-based token refinement is proposed in~\cite{wang2026beyond} to mitigate CLS token bias inherited from pre-training and enhance the modeling of localized forgery cues.

\vspace{.4em}\noindent\textbf{Audio-only methods.}
Early methods~\cite{wu2017asvspoof} rely on hand-crafted features (e.g., MFCC, LFCC) to capture synthesis artifacts, but generalize poorly to diverse channels and unseen generators. With the rise of deep learning, audio spoofing detectors have shifted toward end-to-end neural architectures. A representative baseline is AASIST~\cite{aasist}, which operates directly on raw waveforms using a Sinc-based convolutional front-end and residual blocks, followed by spectral–temporal attention for classification. To further improve robustness, recent approaches~\cite{wav2vec+aasist,iwax,wpt} incorporate self-supervised learning (SSL) representations~\cite{wav2vec2.0,xlsr,avhubert}. By leveraging large-scale pretraining, these methods provide more transferable acoustic features and have become the dominant paradigm in modern audio AIGC detection.

However, despite achieving remarkable success in uni-modal AIGC detection, these methods are inherently limited by their reliance on a single modality and fail to fully leverage the comprehensive cues available across modalities.

\subsection{Audio-Visual AIGC Detection}
Recent studies have increasingly explored leveraging both audio and visual modalities for more reliable AIGC detection.
AVoid-DF \cite{avoid-df} detects multi-modal deepfakes by exploiting audio-visual inconsistency. 
AVGraph \cite{AVgraph} constructs heterogeneous audio-visual graphs to achieve fine-grained forgery classification.
Several audio-visual methods also build upon pre-trained models. The authors of AVFF~\cite{avff} first pre-train the model to learn audio-visual correspondences in a self-supervised stage, followed by supervised deepfake classification.
AVPrompt~\cite{avprompt} fine-tunes CLIP~\cite{clip} and Whisper~\cite{whisper} for deepfake detection via prompt learning.  HAVIC~\cite{havic} detects AIGC based on holistic audio-visual intrinsic coherence priors in the pre-trained models. 
In addition, unsupervised methods, including AVAD~\cite{avad}, SpeechForensics~\cite{speechforensics}, and AVH-Align~\cite{avh-align}, detect forgeries by measuring the matching degree between audio and video, without requiring ground-truth labels.

However, these methods are typically built upon the assumption of strong audio-visual correspondence in authentic content. In general videos, audio may be weakly related, or even unrelated to the visual content, making audio-visual inconsistency a natural characteristic rather than a sign of manipulation.
Consequently, the effectiveness of existing multi-modal detectors may be limited in general AIGC scenarios.
\section{Methodology}
\label{sec:method}
\subsection{Preprocessing}
We first split the MVAD dataset~\cite{mvad} into an audio dataset $\mathcal{D}_a = \{(x_i^a, y_i^a)\}$ and an image dataset $\mathcal{D}_v = \{({x}_{i}^v, y_i^v)\}$. For $\mathcal{D}_a$, the waveform ${x}_i^a$ is directly extracted from each sample, and $y_i^a \in \{0, 1\}$ indicates whether the audio is real (0) or fake (1). For $\mathcal{D}_v$, to balance classes, we uniformly sample 8 images from real videos and 16 images from fake videos to construct the dataset. All sampled images inherit the corresponding video-level label.

\subsection{Visual AIGC Detector}
\label{sec:video}
\noindent\textbf{Feature Extraction.}
Given an input image, we employ DINOv3~\cite{dinov3} as the backbone to extract visual features, yielding a CLS token ${c} \in \mathbb{R}^{D}$ and a set of patch tokens $\{p_i\}_{i=1}^N \in \mathbb{R}^{N \times D}$, where $N$ is the number of patches and $D$ is the feature dimension. The CLS token captures holistic information of the image, while the patch tokens preserve fine-grained spatial local details.

\vspace{.4em}\noindent\textbf{Multi-Granularity Representation Construction.}
AIGC visual cues may emerge at multiple spatial granularities, from subtle local texture artifacts and regional inconsistencies to global semantic anomalies. 
To capture such multi-scale cues, we construct a multi-granularity representation consisting of global, patch-level, and segment-level representations.

Since the CLS token $c_v$ serves as a global summary by aggregating information from all image patches through self-attention~\cite{vit}, we directly adopt it as the global representation.

For patch-level representation, we first employ a two-layer MLP as a patch authenticity scoring network \(Q_p(\cdot)\) to estimate forgery logits $\{l_i^p\}_{i=1}^N$ for each patch token. The logits are normalized via a softmax function with temperature $\tau$:
\begin{equation}
    w_i^p = \frac{\exp(l_i^p / \tau)}{\sum_{j=1}^{N}\exp(l_j^p / \tau)},
\end{equation}

\noindent and used as aggregation weights to compute a weighted sum of patch tokens, producing the patch-level representation:
\begin{equation}
    p_{agg} = \sum_{i=1}^{N} w_i^p p_i .
\end{equation}

For segment-level representation, inspired by~\cite{insid3}, we partition the patch tokens into \(K\) disjoint spatial segments
\(\mathcal{S}=\{S_1,\ldots,S_K\}\)
via agglomerative clustering~\cite{agglo} based on feature cosine similarity. The clustering procedure progressively merges similar neighboring patches in a bottom-up manner, yielding
\begin{equation}
\bigcup_{k=1}^{K} S_k = \{p_i\}_{i=1}^N, \qquad S_i \cap S_j = \emptyset \quad \forall i \neq j.
\end{equation}

For each segment \(S_k\), a segment token \(s_k\) is obtained by average pooling the patch tokens within the segment:
\begin{equation}
s_k=\frac{1}{|S_k|}\sum_{p_i\in S_k} p_i.
\end{equation}

A two-layer MLP is then employed as a segment authenticity scoring network \(Q_s(\cdot)\) to estimate forgery logits \(\{l_k^s\}_{k=1}^{K}\) for each segment token. The logits are normalized using a softmax function with temperature \(\tau\):
\begin{equation}
w_k^s=
\frac{\exp(l_k^s/\tau)}
{\sum_{j=1}^{K}\exp(l_j^s/\tau)}.
\end{equation}

Similarly, the normalized logits are then used as aggregation weights to compute the segment-level representation:
\begin{equation}
s_{agg}
=
\sum_{k=1}^{K}
w_k^s s_k.
\end{equation}

\vspace{.4em}\noindent\textbf{Weak Supervision via Multiple Instance Learning.}
Since only image-level labels are available, the patch and segment authenticity scoring networks, \(Q_p(\cdot)\) and \(Q_s(\cdot)\), lack explicit supervision at the patch and segment levels. We therefore adopt a Multiple Instance Learning strategy to provide weak supervision for both networks. Following the assumption that manipulated images typically contain a portion of highly suspicious patches or segments, 
the top-$k$ forgery logits are aggregated to obtain image-level predictions:
\begin{equation}
l^p_{\text{topk}}
=
\frac{1}{|\mathcal{T}_p|}
\sum_{i \in \mathcal{T}_p}
l_i^p, \quad
l^s_{\text{topk}}
=
\frac{1}{|\mathcal{T}_s|}
\sum_{i \in \mathcal{T}_s}
l_i^s,
\end{equation}
where $\mathcal{T}_*$ denotes the set of the top-$k$ instances ranked by logits. The corresponding Multiple Instance Learning loss is
\begin{equation}
\mathcal{L}_{\text{MIL}}
=
\mathcal{L}_{\text{focal}}
\big(y, \sigma(l_{\text{topk}}^p)\big)
+
\mathcal{L}_{\text{focal}}
\big(y, \sigma(l_{\text{topk}}^s)\big),
\end{equation}
where \(y \in \{0,1\}\) denotes the image-level ground-truth label, $\mathcal{L}_{\text{focal}}$ denotes the focal loss~\cite{focal}, and $\sigma(\cdot)$ denotes the sigmoid function.

To further encourage the model to distinguish suspicious instances from the remaining ones, we introduce a margin loss. Specifically, we compute the forgery logits gap between the top-$k$ instances and the remaining instances:
\begin{equation}
d_p
=
\frac{1}{|\mathcal{T}_p|}
\sum_{i \in \mathcal{T}_p}
l_i^p
-
\frac{1}{N-|\mathcal{T}_p|}
\sum_{i \notin \mathcal{T}_p}
l_i^p.
\end{equation}

\begin{equation}
d_s
=
\frac{1}{|\mathcal{T}_s|}
\sum_{i \in \mathcal{T}_s}
l_i^s
-
\frac{1}{K-|\mathcal{T}_s|}
\sum_{i \notin \mathcal{T}_s}
l_i^s.
\end{equation}

The margin loss is formulated as
\begin{equation}
\mathcal{L}_{\text{margin}}
=
\max(m-d_p,0) + \max(m-d_s,0),
\end{equation}
where $m$ denotes a predefined margin.

The overall weak supervision loss is defined as
\begin{equation}
\mathcal{L}_{\text{weak}}
=
\mathcal{L}_{\text{MIL}}
+
\mathcal{L}_{\text{margin}}.
\end{equation}

\vspace{.4em}
\noindent\textbf{Classifier Network.}
We employ a three-layer MLP that serves as the main classifier, taking the concatenated multi-granularity representation $c_v$, $p_{agg}$, and $s_{agg}$
as input to classify whether an image is real or fake. To prevent the classifier from relying excessively on a particular representation level, we introduce three auxiliary classifiers that operate on $c_v$, $p_{agg}$, and $s_{agg}$, respectively. All classifiers are trained using the focal loss~\cite{focal} with the label $y_v$ as supervision, while only the main classifier is used during inference. The supervised loss is defined as the sum of the focal losses
from the main classifier and the three auxiliary classifiers:
\begin{equation}
\mathcal{L}_{\text{sup}}
=
\mathcal{L}_{\text{main}}
+
\mathcal{L}_{\text{global}}
+ \mathcal{L}_{\text{patch}}
+ \mathcal{L}_{\text{segment}}.
\end{equation}

\begin{table*}[t]
\centering
\caption{Perturbation methods used in the Degraded-Original Consistency Learning.}
\vspace{-0.5em}
\label{tab:perturbations}
\begin{tabular}{ll}
\toprule
\textbf{Category} & \textbf{Methods} \\
\midrule
Geometric Transformations & Horizontal Flip, Image Rotation, Image Translation, Image Resizing \\
Noise & Gaussian Noise, Impulse Noise, Speckle Noise, Poisson Noise \\
Blur & Gaussian Blur, Lens Defocus Blur, Mean Blur \\
Color Distortion & Saturation Change, Grayscale, Quantization \\
Photometric Transformations & Brighten, Darken, Contrast Change \\
Compression & JPEG Compression \\
\bottomrule
\end{tabular}
\end{table*}

\vspace{.4em}\noindent\textbf{Degraded-Original Consistency Learning.}
To improve the robustness of the detector, we introduce diverse image perturbations as data augmentation, including noise, blur, geometric transformations, and color jittering, to simulate real-world image degradations. Based on these augmented views, to encourage the model to learn degradation-robust representation, we further introduce a Degraded-Original Consistency Learning (DOCL) strategy. For each original input image $x^{ori}_v$, we generate a degraded counterpart $x^{deg}_v$ using the perturbation pipeline. $x^{ori}_v$ and $x^{deg}_v$ are separately fed into the backbone. A consistency loss $
\mathcal{L}_\text{con}$ is employed to enforce multi-granularity representation consistency between the original and degraded images:
\begin{equation}
\mathcal{L}_{\text{con}}
=
\sum_{z \in \{c_v, p_{{agg}}, s_{{agg}}\}}
\left(
1 - \text{sim}\big(\text{sg}[z^{ori}], z^{deg}\big)
\right),
\end{equation}
where $\text{sim}(\cdot, \cdot)$ denotes the cosine similarity, and sg[$\cdot$] represents the stop-gradient operation, meaning that gradients from $\mathcal{L}_\text{con}$ are only propagated through the degraded representations, while the representations of the original image are treated as fixed targets.

Finally, the overall visual loss function is defined as:
\begin{equation}
\mathcal{L}_v
=
\mathcal{L}_{\text{sup}}^{ori}
+
\mathcal{L}_{\text{weak}}^{ori}
+
\mathcal{L}_{\text{sup}}^{deg}
+
\mathcal{L}_{\text{weak}}^{deg}
+
0.05 * \mathcal{L}_{\text{con}}.
\end{equation}

\subsection{Audio AIGC Detector}
\label{sec:audio}
\noindent\textbf{Feature Extraction.} We leverage the audio encoder of PEAV~\cite{PEAV} as the audio backbone.
The input audio waveform is fed into the backbone to extract audio features. The output comprises a CLS token $c_a \in \mathbb{R}^{D}$ and a sequence of frame tokens $\{f_t\}_{t=1}^{T} \in \mathbb{R}^{T \times D}$, where $T$ denotes the number of audio frames. The CLS token captures holistic acoustic properties, while the frame tokens preserve fine-grained temporal-spectral details.


\vspace{.4em}\noindent\textbf{Temporal-Spectral Representation Construction.}
AIGC audio artifacts manifest in both temporal dynamics (e.g., prosody inconsistency and unnatural transitions) and spectral characteristics (e.g., high-frequency artifacts). To capture these multi-aspect cues, inspired by~\cite{aasist}, we design a temporal-spectral dual-branch architecture.

To adaptively modulate discriminative temporal-spectral features, we compute a feature-wise gating map via a gating network \(G(\cdot)\), implemented as a two-layer MLP:

\begin{equation}
{W} = \sigma({G}(\{{f}_t\}_{t=1}^{T})) \in (0,1)^{T \times D},
\end{equation}

\noindent where $\sigma$ denotes the sigmoid function. The gating map is then used to modulate the audio frame token sequence:
\begin{equation}
\{ \tilde{f}_t \}_{t=1}^{T} = \{ f_t  \}_{t=1}^{T} \odot W.
\end{equation}

The temporal branch first refines the modulated frame sequence $ \{\tilde{f}_t\}_{t=1}^{T}$ by modeling temporal dependencies across frames using self-attention~\cite{transformer}, yielding a temporally enhanced feature sequence $F_{\text{temp}}$. Based on $F_{\text{temp}}$, we construct complementary temporal representations that summarize different aspects of the temporal dynamics, including an average token ${f}_{\text{temp}}^{\text{avg}}$ via temporal average pooling, a discriminative token ${f}_{\text{temp}}^{\max}$ via max pooling, and an adaptive token ${f}_{\text{temp}}^{\mu}$ via learnable query-based pooling.

The spectral branch operates on the same modulated features and employs a channel attention~\cite{squeeze} mechanism to emphasize spectral-related patterns. 
This process yields a spectrally enhanced  feature sequence $F_{\text{spec}}$. 
Based on $F_{\text{spec}}$, we derive complementary spectral representations via different pooling strategies in the spectral domain, including an average token $f_{\text{spec}}^{\text{avg}}$, a discriminative token $f_{\text{spec}}^{\max}$, and a learnable query-based token $f_{\text{spec}}^{\mu}$.

To integrate global and local cues, the CLS token $c_a$ attends to the above representations via cross-attention~\cite{transformer}. This process allows the model to selectively aggregate informative signals from both temporal and spectral perspectives, producing a unified audio representation $m$.

\vspace{.4em}\noindent\textbf{Classifier Network.}
The final representation $\mathbf{r}$ is constructed by concatenating the unified audio representation with temporal and spectral representations:
\begin{equation}
\mathbf{r} = \text{Concat}[{m}, {f}_{\text{temp}}^{\text{avg}}, {f}_{\text{temp}}^{\max}, {f}_{\text{temp}}^{\mu}, {f}_{\text{spec}}^{\text{avg}}, {f}_{\text{spec}}^{\max}, {f}_{\text{spec}}^{\mu}].
\end{equation}

We employ a three-layer MLP as the audio classifier, which takes $\mathbf{r}$ as input and predicts whether the input audio is real or fake. The classifier is trained using Focal Loss~\cite{focal} with audio label $y_a$ as supervision.

\subsection{Decision-Level Fusion for Inference}
During inference, the audio and visual detectors independently produce fake probabilities for each input.
For the visual stream, we uniformly sample 16 frames from each video clip. The visual detector produces frame-level logits, which are then averaged to obtain the final video-level logit $\bar{l}_v$. A sigmoid function is applied to obtain the fake probability:
\begin{equation}
p_v^{Fake} = \sigma(\bar{l}_v).
\end{equation}

For the audio stream, we adopt a sliding window strategy with a window size and stride of 3 seconds. The audio waveform is segmented into consecutive windows, and the audio detector produces window-level logits. These logits are then averaged to obtain the final audio-level logit $\bar{l}_a$.  A sigmoid function is applied to obtain the fake probability:
\begin{equation}
p_a^{Fake} = \sigma(\bar{l}_a).
\end{equation}

The corresponding real probabilities are defined as:
\begin{equation}
p_a^{Real} = 1 - p_a^{Fake}, \quad p_v^{Real} = 1 - p_v^{Fake}.
\end{equation}

For binary classification, we adopt a max-based fusion strategy. The final fake probability is defined as:
\begin{equation}
p^{Fake} = \max(p_a^{Fake}, p_v^{Fake}),
\end{equation}
and the real probability is computed as:
\begin{equation}
p^{Real} = 1 - p^{Fake}.
\end{equation}

This design assumes that a sample is fake if either the audio or visual modality exhibits a high fake probability.

For four-class classification, we define a joint audio-visual probability distribution under an independence assumption:
\begin{equation}
\begin{aligned}
p_{RR} &= p_v^{Real} \cdot p_a^{Real}, \\
p_{FF} &= p_v^{Fake} \cdot p_a^{Fake}, \\
p_{FR} &= p_v^{Fake} \cdot p_a^{Real}, \\
p_{RF} &= p_v^{Real} \cdot p_a^{Fake},
\end{aligned}
\end{equation}

\noindent where $p_{RR}$ denotes the case where both modalities are real, and $p_{FF}$ indicates that both are fake. $p_{FR}$ and $p_{RF}$ correspond to the video being fake while the audio is real, and the video being real while the audio is fake, respectively. The final prediction is obtained by selecting the class with the highest joint probability.

\begin{table*}[t]
\centering
\caption{Results of the AIGC Audio-Video Detection Challenge (Top-5 teams). Best result is in \textbf{bold}, and second best is \underline{underlined}.}
\vspace{-0.5em}
\label{tab:challenge-results}
\setlength{\tabcolsep}{4pt}
\renewcommand{\arraystretch}{1.3}
\begin{adjustbox}{width=1\textwidth,center}
\begin{tabular}{clccccccc}
\toprule
\textbf{Rank} &
\textbf{Team} &
\textbf{Final Score} &
\textbf{Binary Score} &
\textbf{Four-Class Score} &
\textbf{Binary AUC} &
\textbf{Binary ACC} &
\textbf{Four-Class F1} &
\textbf{Four-Class AP} 
 \\
\midrule
\textbf{1} & \textbf{HIT VIRLAB (Ours)}     &\textbf{0.8460}&	0.9379	&\underline{0.7847}&	\textbf{0.9617}	&\underline{0.9190}	&\textbf{0.6873}	&\textbf{0.7274}\\
\underline{2} & \underline{ZJSU} & \underline{0.8438}	&\underline{0.9395}&	0.7800&	\underline{0.9490}	&\textbf{0.9217}	&0.6847	&0.7149 \\
3 & VeriLens &0.8426&	0.9288&	\textbf{0.7851}&	0.9288&	0.9106&	\underline{0.6856}&	0.7096\\
4 & AIGVDete & 0.8406&	\textbf{0.9396}&	0.7746&	0.9480&	0.9187&	0.6726&	\underline{0.7191}\\
5 & MVADetection Team & 0.8349& 0.9278&	0.7729&	0.9447&	0.9000&	0.6530&	0.7167\\
\bottomrule
\end{tabular}
\end{adjustbox}
\vspace{-1em}
\end{table*}

\section{Experiments}
\subsection{Experimental Setup}
\noindent\textbf{Dataset and metrics.} 
We evaluate our method on the General AIGC Audio--Video Detection benchmark (DDL-GAV), which is built upon the MVAD dataset~\cite{mvad}. DDL-GAV is a large-scale multimodal benchmark containing over 200,000 video-audio samples covering both realistic and anime visual domains, four content categories (humans, animals, objects, and scenes), and more than 23 state-of-the-art AIGC generation methods. Based on which modalities
are generated by AI, the samples can be categorized into four
types: real video with fake audio (RF), fake video with
real audio (FR), fake video with fake audio (FF), and
real video with real audio (RR, i.e., the authentic samples). 

The benchmark includes two hierarchical tasks: binary forgery detection and four-class modality classification. For binary detection, samples are classified as either real or fake. For four-class classification, each sample belongs to one of the following categories: RF, FR, FF, and RR. Following the official evaluation protocol, model performance is assessed using Area Under the ROC Curve (AUC), Accuracy (ACC), Average Precision (AP), and F1-score. For both binary and four-class classification, the official score is computed as:
\begin{equation}
\text{Score} = 0.2 \times \text{AUC} +
0.3 \times \text{ACC} +
0.3 \times \text{AP} +
0.2 \times \text{F1}.
\end{equation}

The final score is computed as a weighted sum of the binary detection score and the four-class classification score, with weights of 0.4 and 0.6, respectively.

\vspace{.4em}\noindent\textbf{Implementation details.}
For the visual AIGC detector, we employ the DINOv3 ViT-L/16~\cite{dinov3} with feature dimension $D=1024$ as the backbone, and apply LoRA adaptation~\cite{lora} (rank 32, alpha 16) for parameter-efficient fine-tuning. 
We train the visual AIGC detector using the AdamW optimizer with a weight decay of 5e-2 and a learning rate of 1e-4, decayed using a cosine schedule.
We train for 20 epochs with a linear warmup for 2 epochs using eight NVIDIA L20 GPUs with a total batch size of 112 and a gradient accumulation interval of 16. The softmax temperature parameter $\tau$ is fixed at 0.07. The predefined margin used in the margin loss is fixed at 0.6. The top-k ratios are set to 0.05 for patches and 0.1 for segments. For agglomerative clustering, we use average linkage with a cosine similarity threshold of 0.9. Data augmentations used for Degraded-Original Consistency Learning are shown in Table~\ref{tab:perturbations}.
For the audio AIGC detector, we employ the audio encoder of PEAV-base~\cite{PEAV} with feature dimension $D=1024$ as the backbone, and apply LoRA adaptation~\cite{lora} (rank 32, alpha 64) for parameter-efficient fine-tuning. We train the audio AIGC detector using the AdamW optimizer with a weight decay of 5e-2 and a learning rate of 1e-4, decayed using a cosine schedule.
We train for 20 epochs with a linear warmup for 2 epochs using four NVIDIA L20 GPUs with a total batch size of 512.

\subsection{Competition Results}
Table~\ref{tab:challenge-results} reports the top-5 results on the General AIGC Audio-Video Detection Challenge leaderboard\footnote{\url{https://www.codabench.org/competitions/15769/\#/results-tab}}. Our team, HIT VIRLAB, achieves the best overall performance with a final score of 0.8460, outperforming all competing teams. 

Beyond the overall ranking, our method demonstrates consistently strong performance across both binary and four-class classification tasks. In binary classification, we achieve the highest AUC of 0.9617, significantly surpassing the second-best result (+1.27\%), while maintaining competitive accuracy. For the more challenging four-class classification, our method attains the best F1 score of 0.6873 and AP score of 0.7274, indicating superior capability in fine-grained modality-aware discrimination.

It is worth noting that AIGVDete achieves the highest binary accuracy, while VeriLens obtains slightly better performance in terms of four-class F1 score. Nevertheless, our approach maintains a more balanced performance across all evaluation metrics, leading to the highest overall ranking on the leaderboard.

\subsection{Comparison with AV Deepfake Detectors}
We further compare our DAV-Det with recent audio-visual methods on the FakeAVCeleb dataset~\cite{fakeavceleb}. For fair comparisons, we follow the same data preprocessing and split protocol in~\cite{avoid-df,avff,pia,fovb}, using 70\% of the FakeAVCeleb samples for training and validation, and the remaining 30\% for testing. Facial regions are cropped from video frames using FaceX-Zoo~\cite{facexzoo} to mitigate background interference. We report binary classification ACC and AUC as evaluation metrics, consistent with prior works for fair comparisons. As shown in Table~\ref{tab:fakeavceleb}, DAV-Det achieves superior performance compared with existing audio-visual deepfake detectors, reaching 0.998 ACC and 0.999 AUC. Although FakeAVCeleb mainly contains human-centric deepfakes rather than general AIGC-generated videos, these results demonstrate the strong detection capability of DAV-Det on conventional human-centric audio-visual deepfake scenarios, benefiting from its effective modality-specific detectors and modality decoupling strategy.
\begin{table}[H]
    \centering
    \vspace{-0.5em}
    \caption{Comparison with recent audio-visual methods on FakeAVCeleb.
    Best result is in \textbf{bold}, and second best is \underline{underlined}.
    }
    \vspace{-0.2em}
    \label{tab:fakeavceleb}
    \renewcommand{\arraystretch}{0.95}
    \setlength{\heavyrulewidth}{0.05em}
    \setlength{\lightrulewidth}{0.03em}
    {\fontsize{4}{5}\selectfont
    \begin{adjustbox}{width=0.95\linewidth,center}
    \begin{tabular}{l c *{2}{c}}
    \toprule
    \textbf{Method} &  \textbf{ACC} & \textbf{AUC}   \\
    \midrule
    VFD~\cite{vfd} &0.815 &0.861 \\
    AVoiD-DF \cite{avoid-df}  & {0.837} & {0.892}  \\
    MCL~\cite{mcl} & 0.860 & 0.896 \\
    MRDF-CE \cite{mrdf} & 0.941 & 0.924\\
    AVFF \cite{avff}  & 0.986 & 0.991 \\
    PIA \cite{pia} & \underline{0.987} & \underline{0.998} \\
    FoVB~\cite{fovb}  & 0.985 & 0.997 \\
    \midrule
    \textbf{DAV-Det (Ours)}  & \textbf{0.998} & \textbf{0.999} \\
    \bottomrule
    \end{tabular}
    \end{adjustbox}
    }
\end{table}

\subsection{Ablation Study}
In this section, we conduct ablation studies to evaluate the effectiveness of each component in DAV-Det. As the ground-truth labels of the official DDL-GAV test set are unavailable, we first randomly sample a subset from the original training set and then divide it into training and validation subsets with a ratio of 9:1. The model is trained on the training subset, and the ablation studies are conducted on the validation subset. For evaluating the audio and visual detectors separately, a video is considered fake only when the corresponding modality is manipulated, i.e., audio manipulation for the audio detector and visual manipulation for the visual detector.

\vspace{.4em}\noindent\textbf{Ablation on Visual Detector.}
We first analyze the effectiveness of the multi-granularity design in the visual detector. The baseline only uses the global branch for classification. We then
introduce patch-level and segment-level branches to enrich multi-granularity representations. As shown in Tab.~\ref{tab:multigranularity}, introducing multi-granularity branches brings progressive improvements over the global-only baseline, with the full model achieving the best results, demonstrating the effectiveness of multi-granularity representation for deepfake detection.

\begin{table}[H]
\centering
\vspace{-0.5em}
\caption{Effectiveness of the multi-granularity design.}
\label{tab:multigranularity}
\vspace{-0.5em}
\setlength{\heavyrulewidth}{0.05em}
\setlength{\lightrulewidth}{0.04em}
{\fontsize{6}{7}\selectfont
\setlength{\tabcolsep}{4pt}
\renewcommand{\arraystretch}{1.10}
\begin{adjustbox}{width=0.98\linewidth,center}
\begin{tabular}{ccccccc}
\toprule
\textbf{Global} & \textbf{Patch} & \textbf{Segment} & \textbf{ACC} & \textbf{AP} & \textbf{AUC}& \textbf{F1}\\
\midrule
\textcolor{green}{\cmark} & \textcolor{red}{\xmark} & \textcolor{red}{\xmark} & 0.9742 & 0.9695 &0.9895 &0.9181 \\
\textcolor{green}{\cmark} & \textcolor{green}{\cmark} & \textcolor{red}{\xmark} & 0.9757 & 0.9706 &0.9916 &0.9228 \\
\midrule
\textcolor{green}{\cmark} & \textcolor{green}{\cmark} & \textcolor{green}{\cmark} & \textbf{0.9836} & \textbf{0.9879} & \textbf{0.9969} & \textbf{0.9482}\\
\bottomrule
\end{tabular}
\end{adjustbox}
}

\end{table}

In addition, we study the contribution of the weak supervision loss, the auxiliary classification head, and the DOCL strategy.  As shown in Tab.~\ref{tab:component_ablation}, removing the weakly-supervised loss leads to a performance drop, indicating that it provides effective weak supervision for the patch and segment authenticity scoring networks. Similarly, removing the auxiliary classifiers degrades performance, suggesting their effectiveness in facilitating multi-granularity representations for capturing deepfake cues. Removing the DOCL strategy also results in performance degradation, demonstrating its effectiveness in improving robustness to input perturbations and maintaining multi-granularity consistency across degraded and original views of the same image.

\begin{table}[H]
\centering
\vspace{-0.5em}
\caption{Ablation study on individual strategies in visual detector.}
\vspace{-0.5em}
\label{tab:component_ablation}
\setlength{\heavyrulewidth}{0.05em}
\setlength{\lightrulewidth}{0.04em}
{\fontsize{6}{7}\selectfont
\setlength{\tabcolsep}{4pt}
\renewcommand{\arraystretch}{1.10}

\begin{adjustbox}{width=0.98\linewidth,center}
\begin{tabular}{lcccc}
\toprule
\textbf{Method} & \textbf{ACC} & \textbf{AP} & \textbf{AUC}& \textbf{F1} \\
\midrule
w/o weak supervision loss & 0.9764 & 0.9773 & 0.9920 & 0.9252  \\
w/o auxiliary classifiers   & 0.9803 &0.9856 & 0.9951 & 0.9437 \\
w/o DOCL strategy     & 0.9750 & 0.9749 & 0.9925 &0.9210    \\
\midrule
\textbf{Ours} & \textbf{0.9836} & \textbf{0.9879} & \textbf{0.9969} & \textbf{0.9482} \\
\bottomrule
\end{tabular}
\end{adjustbox}
}

\vspace{-1em}
\end{table}

\vspace{.4em}\noindent\textbf{Ablation on Audio Detector.}
We further investigate the effectiveness of the proposed temporal-spectral dual-branch design in the audio detector. As shown in Tab.~\ref{tab:temporal-spectral}, removing both branches (i.e., only using the CLS token for classification) causes a noticeable performance degradation, indicating the importance of explicitly modeling temporal-spectral audio representations for audio AIGC detection. Introducing the temporal branch improves the detection performance over the baseline, demonstrating that temporal dynamics provide valuable forgery-related cues. Similarly, incorporating the spectral branch further enhances the detection capability, indicating that spectral information also contains discriminative cues for audio AIGC detection. Finally, combining both temporal and spectral branches achieves the best performance.
These results verify that temporal and spectral information provide complementary evidence, and their joint modeling enables more effective audio AIGC detection.

\begin{table}[H]
\centering
\vspace{-0.5em}
\caption{Effectiveness of the temporal-spectral design.}
\label{tab:temporal-spectral}
\vspace{-0.5em}
\setlength{\heavyrulewidth}{0.04em}
\setlength{\lightrulewidth}{0.03em}
{\fontsize{4}{5}\selectfont
\setlength{\tabcolsep}{4pt}
\renewcommand{\arraystretch}{0.95}

\begin{adjustbox}{width=0.98\linewidth,center}
\begin{tabular}{cccccc}
\toprule
 \textbf{Temporal} & \textbf{Spectral} & \textbf{ACC} & \textbf{AP} & \textbf{AUC} & \textbf{F1} \\
\midrule
\textcolor{red}{\xmark} & \textcolor{red}{\xmark} & 0.9700 & 0.9759 & 0.9914 & 0.9302 \\
\textcolor{green}{\cmark} & \textcolor{red}{\xmark} & 0.9736 & 	0.9783 & 0.9917 & 0.9387  \\
\textcolor{red}{\xmark} & \textcolor{green}{\cmark} & 0.9724 & 0.9773 & 0.9904 & 0.9358 \\
\midrule
\textcolor{green}{\cmark} & \textcolor{green}{\cmark} & \textbf{0.9791} & \textbf{0.9829}& \textbf{0.9942} &\textbf{0.9435} \\
\bottomrule
\end{tabular}
\end{adjustbox}
}
\end{table}

\section{Conclusion, Limitations, and Future Work}
In this paper, we study general AIGC audio-video detection and empirically demonstrate that the assumption of audio-visual content correspondence, which underlies most existing multi-modal deepfake detectors, does not consistently hold in general AIGC scenarios.
To address this, we propose DAV-Det, a decoupled audio-visual detection system  that models each modality independently and performs decision-level fusion. The visual detector leverages multi-granularity representations at global, patch, and segment levels to capture forgery cues at different spatial levels, while the audio detector exploits both temporal and spectral irregularities via a gated temporal-spectral dual-branch architecture to model acoustic artifacts. 
Our method achieves first place in the General AIGC Audio-Video Detection Challenge at the IJCAI-ECAI 2026 DDL 2.0 Workshop, with a final score of 0.8460.

\vspace{.4em}\noindent\textbf{Limitations.} 
Despite the effectiveness of DAV-Det, several limitations remain. First, the visual detector does not explicitly model fine-grained temporal dependencies, such as motion inconsistencies across frames. Second, the decision-level fusion strategy relies on heuristic rules, including max-based fusion for binary classification and independence assumptions for four-class prediction. These choices may not fully leverage the potential of adaptive decision-level fusion.

\vspace{.4em}\noindent\textbf{Future Work.} 
Future work will explore temporal visual modeling to capture dynamic forgery traces across frames and develop more principled audio-visual decision-level fusion strategies, such as adaptive weighting and learnable calibration, to further improve audio-visual AIGC detection.

\section*{Acknowledgements}
This work is funded by the National Key R\&D Program of China (No. 2025YFC3811300), the National Natural Science Foundation of China (Grant No. 62376070 and 62076195), the Fundamental Research Funds for the Central Universities (AUGA5710028726), and the China Postdoctoral Science Foundation (No. 2026M794773).

\bibliographystyle{named}
\bibliography{ijcai26}

@inproceedings{havic,
  title={Leave No Stone Unturned: Uncovering Holistic Audio-Visual Intrinsic Coherence for Deepfake Detection},
  author={Peng, Jielun and Wang, Yabin and Li, Yaqi and Kong, Long and Hong, Xiaopeng},
  booktitle={Proceedings of the IEEE/CVF Conference on Computer Vision and Pattern Recognition},
  pages={6655--6666},
  year={2026}
}

@inproceedings{PEAV,
  title={Pushing the frontier of audiovisual perception with large-scale multimodal correspondence learning},
  author={Vyas, Apoorv and Chang, Heng-Jui and Yang, Cheng-Fu and Huang, Po-Yao and Gao, Luya and Richter, Julius and Chen, Sanyuan and Le, Matthew and Doll{\'a}r, Piotr and Feichtenhofer, Christoph and others},
  booktitle={Proceedings of the IEEE/CVF Conference on Computer Vision and Pattern Recognition},
  pages={30172--30182},
  year={2026}
}

@article{mvad,
  title={MVAD: A Comprehensive Multimodal Video-Audio Dataset for AIGC Detection},
  author={Hu, Mengxue and Diao, Yunfeng and Miao, Changtao and Li, Jianshu and Li, Zhe and Zhou, Joey Tianyi},
  journal={arXiv preprint arXiv:2512.00336},
  year={2025}
}

@article{poundnet,
  title={Penny-wise and pound-foolish in ai-generated image detection},
  author={Wang, Yabin and Huang, Zhiwu and Su, Zhou and Prugel-Bennett, Adam and Hong, Xiaopeng},
  journal={IEEE Transactions on Pattern Analysis and Machine Intelligence},
  year={2026},
  publisher={IEEE}
}

@inproceedings{opensdi,
  title={Opensdi: Spotting diffusion-generated images in the open world},
  author={Wang, Yabin and Huang, Zhiwu and Hong, Xiaopeng},
  booktitle={Proceedings of the IEEE/CVF Conference on Computer Vision and Pattern Recognition},
  pages={4291--4301},
  year={2025}
}

@inproceedings{aasist,
  title={Aasist: Audio anti-spoofing using integrated spectro-temporal graph attention networks},
  author={Jung, Jee-weon and Heo, Hee-Soo and Tak, Hemlata and Shim, Hye-jin and Chung, Joon Son and Lee, Bong-Jin and Yu, Ha-Jin and Evans, Nicholas},
  booktitle={ICASSP 2022-2022 IEEE international conference on acoustics, speech and signal processing (ICASSP)},
  pages={6367--6371},
  year={2022},
  organization={IEEE}
}

@article{rawbmamba,
  title={Rawbmamba: End-to-end bidirectional state space model for audio deepfake detection},
  author={Chen, Yujie and Yi, Jiangyan and Xue, Jun and Wang, Chenglong and Zhang, Xiaohui and Dong, Shunbo and Zeng, Siding and Tao, Jianhua and Zhao, Lv and Fan, Cunhang},
  journal={arXiv preprint arXiv:2406.06086},
  year={2024}
}

@inproceedings{allm4add,
  title={Allm4add: Unlocking the capabilities of audio large language models for audio deepfake detection},
  author={Gu, Hao and Yi, Jiangyan and Wang, Chenglong and Tao, Jianhua and Lian, Zheng and He, Jiayi and Ren, Yong and Chen, Yujie and Wen, Zhengqi},
  booktitle={Proceedings of the 33rd ACM International Conference on Multimedia},
  pages={11736--11745},
  year={2025}
}

@inproceedings{avprompt,
  title={Multi-modal Deepfake Detection via Multi-task Audio-Visual Prompt Learning},
  author={Miao, Hui and Guo, Yuanfang and Liu, Zeming and Wang, Yunhong},
  booktitle={Proceedings of the AAAI Conference on Artificial Intelligence},
  volume={39},
  number={1},
  pages={612--621},
  year={2025}
}

@inproceedings{avad,
  title={Self-Supervised Video Forensics by Audio-Visual Anomaly Detection},
  author={Feng, Chao and Chen, Ziyang and Owens, Andrew},
  booktitle={Proceedings of the IEEE/CVF Conference on Computer Vision and Pattern Recognition},
  pages={10491--10503},
  year={2023}
}

@article{speechforensics,
  title={SpeechForensics: Audio-visual speech representation learning for face forgery detection},
  author={Liang, Yachao and Yu, Min and Li, Gang and Jiang, Jianguo and Li, Boquan and Yu, Feng and Zhang, Ning and Meng, Xiang and Huang, Weiqing},
  journal={Advances in Neural Information Processing Systems},
  volume={37},
  pages={86124--86144},
  year={2024}
}

@article{fakeavceleb,
  title={FakeAVCeleb: A novel audio-video multimodal deepfake dataset},
  author={Khalid, Hasam and Tariq, Shahroz and Kim, Minha and Woo, Simon S},
  journal={arXiv preprint arXiv:2108.05080},
  year={2021}
}

@article{demamba,
  title={Demamba: Ai-generated video detection on million-scale genvideo benchmark},
  author={Chen, Haoxing and Hong, Yan and Huang, Zizheng and Xu, Zhuoer and Gu, Zhangxuan and Li, Yaohui and Lan, Jun and Zhu, Huijia and Zhang, Jianfu and Wang, Weiqiang and others},
  journal={Science China Information Sciences},
  volume={69},
  number={6},
  pages={162103},
  year={2026},
  publisher={Springer}
}

@article{wang2025benchmark,
  title={Benchmark dataset and framework for continual AI-generated image detection},
  author={Wang, Yabin and Hong, Xiaopeng and Huang, Zhiwu},
  journal={Journal of Image and Graphics},
  volume={30},
  number={11},
  pages={3438--3450},
  year={2025},
  publisher={Editorial and Publishing Board of JIG}
}

@article{wang2024linguistic,
  title={Linguistic profiling of deepfakes: An open database for next-generation deepfake detection},
  author={Wang, Yabin and Huang, Zhiwu and Ma, Zhiheng and Hong, Xiaopeng},
  journal={arXiv preprint arXiv:2401.02335},
  year={2024}
}

@article{dinov3,
  title={Dinov3},
  author={Sim{\'e}oni, Oriane and Vo, Huy V and Seitzer, Maximilian and Baldassarre, Federico and Oquab, Maxime and Jose, Cijo and Khalidov, Vasil and Szafraniec, Marc and Yi, Seungeun and Ramamonjisoa, Micha{\"e}l and others},
  journal={arXiv preprint arXiv:2508.10104},
  year={2025}
}

@inproceedings{insid3,
  title={INSID3: Training-Free In-Context Segmentation with DINOv3},
  author={Cuttano, Claudia and Trivigno, Gabriele and Reich, Christoph and Cremers, Daniel and Masone, Carlo and Roth, Stefan},
  booktitle={Proceedings of the IEEE/CVF Conference on Computer Vision and Pattern Recognition},
  pages={21638--21648},
  year={2026}
}

@article{agglo,
  title={Modern hierarchical, agglomerative clustering algorithms},
  author={M{\"u}llner, Daniel},
  journal={arXiv preprint arXiv:1109.2378},
  year={2011}
}

@inproceedings{focal,
  title={Focal loss for dense object detection},
  author={Lin, Tsung-Yi and Goyal, Priya and Girshick, Ross and He, Kaiming and Doll{\'a}r, Piotr},
  booktitle={Proceedings of the IEEE international conference on computer vision},
  pages={2980--2988},
  year={2017}
}

@String(ICASSP=	{ICASSP})

@String(ICLR = {Int. Conf. Learn. Represent.})

@String(AAAI = {AAAI})

@String(ICLR  = {ICLR})

@inproceedings{avff,
  title={Avff: Audio-visual feature fusion for video deepfake detection},
  author={Oorloff, Trevine and Koppisetti, Surya and Bonettini, Nicol{\`o} and Solanki, Divyaraj and Colman, Ben and Yacoob, Yaser and Shahriyari, Ali and Bharaj, Gaurav},
  booktitle={Proceedings of the IEEE/CVF Conference on Computer Vision and Pattern Recognition},
  pages={27102--27112},
  year={2024}
}

@article{AVgraph,
  title={Fine-grained multimodal deepfake classification via heterogeneous graphs},
  author={Yin, Qilin and Lu, Wei and Cao, Xiaochun and Luo, Xiangyang and Zhou, Yicong and Huang, Jiwu},
  journal={International Journal of Computer Vision},
  volume={132},
  number={11},
  pages={5255--5269},
  year={2024},
  publisher={Springer}
}

@article{vit,
  title={An image is worth 16x16 words: Transformers for image recognition at scale},
  author={Dosovitskiy, Alexey and Beyer, Lucas and Kolesnikov, Alexander and Weissenborn, Dirk and Zhai, Xiaohua and Unterthiner, Thomas and Dehghani, Mostafa and Minderer, Matthias and Heigold, Georg and Gelly, Sylvain and others},
  journal={arXiv preprint arXiv:2010.11929},
  year={2020}
}

@inproceedings{facexzoo,
  title={Facex-zoo: A pytorch toolbox for face recognition},
  author={Wang, Jun and Liu, Yinglu and Hu, Yibo and Shi, Hailin and Mei, Tao},
  booktitle={Proceedings of the 29th ACM international conference on Multimedia},
  pages={3779--3782},
  year={2021}
}

@article{transformer,
  title={Attention is all you need},
  author={Vaswani, Ashish and Shazeer, Noam and Parmar, Niki and Uszkoreit, Jakob and Jones, Llion and Gomez, Aidan N and Kaiser, {\L}ukasz and Polosukhin, Illia},
  journal={Advances in neural information processing systems},
  volume={30},
  year={2017}
}

@inproceedings{mrdf,
  title={Cross-modality and within-modality regularization for audio-visual deepfake detection},
  author={Zou, Heqing and Shen, Meng and Hu, Yuchen and Chen, Chen and Chng, Eng Siong and Rajan, Deepu},
  booktitle={ICASSP 2024-2024 IEEE International Conference on Acoustics, Speech and Signal Processing (ICASSP)},
  pages={4900--4904},
  year={2024},
  organization={IEEE}
}

@inproceedings{lip-forensics,
  title={Lips don't lie: A generalisable and robust approach to face forgery detection},
  author={Haliassos, Alexandros and Vougioukas, Konstantinos and Petridis, Stavros and Pantic, Maja},
  booktitle={Proceedings of the IEEE/CVF conference on computer vision and pattern recognition},
  pages={5039--5049},
  year={2021}
}

@article{avoid-df,
  title={Avoid-df: Audio-visual joint learning for detecting deepfake},
  author={Yang, Wenyuan and Zhou, Xiaoyu and Chen, Zhikai and Guo, Bofei and Ba, Zhongjie and Xia, Zhihua and Cao, Xiaochun and Ren, Kui},
  journal={IEEE Transactions on Information Forensics and Security},
  volume={18},
  pages={2015--2029},
  year={2023},
  publisher={IEEE}
}

@article{vfd,
  title={Voice-face homogeneity tells deepfake},
  author={Cheng, Harry and Guo, Yangyang and Wang, Tianyi and Li, Qi and Chang, Xiaojun and Nie, Liqiang},
  journal={ACM Transactions on Multimedia Computing, Communications and Applications},
  volume={20},
  number={3},
  pages={1--22},
  year={2023},
  publisher={ACM New York, NY}
}

@inproceedings{avh-align,
  title={Circumventing shortcuts in audio-visual deepfake detection datasets with unsupervised learning},
  author={Smeu, Stefan and Boldisor, Dragos-Alexandru and Oneata, Dan and Oneata, Elisabeta},
  booktitle={Proceedings of the Computer Vision and Pattern Recognition Conference},
  pages={18815--18825},
  year={2025}
}

@inproceedings{pia,
  title={PIA: Deepfake Detection Using Phoneme-Temporal and Identity-Dynamic Analysis},
  author={Datta, Soumyya Kanti and Ranga, Tanvi and Sun, Chengzhe and Lyu, Siwei},
  booktitle={Proceedings of the IEEE/CVF International Conference on Computer Vision},
  pages={1596--1606},
  year={2025}
}

@inproceedings{Multi_attentional_deepfake_detection,
  title={Multi-attentional deepfake detection},
  author={Zhao, Hanqing and Zhou, Wenbo and Chen, Dongdong and Wei, Tianyi and Zhang, Weiming and Yu, Nenghai},
  booktitle={Proceedings of the IEEE/CVF conference on computer vision and pattern recognition},
  pages={2185--2194},
  year={2021}
}

@article{avhubert,
  title={Learning audio-visual speech representation by masked multimodal cluster prediction},
  author={Shi, Bowen and Hsu, Wei-Ning and Lakhotia, Kushal and Mohamed, Abdelrahman},
  journal={arXiv preprint arXiv:2201.02184},
  year={2022}
}

@inproceedings{clip,
  title={Learning transferable visual models from natural language supervision},
  author={Radford, Alec and Kim, Jong Wook and Hallacy, Chris and Ramesh, Aditya and Goh, Gabriel and Agarwal, Sandhini and Sastry, Girish and Askell, Amanda and Mishkin, Pamela and Clark, Jack and others},
  booktitle={International conference on machine learning},
  pages={8748--8763},
  year={2021},
  organization={PmLR}
}

@inproceedings{whisper,
  title={Robust speech recognition via large-scale weak supervision},
  author={Radford, Alec and Kim, Jong Wook and Xu, Tao and Brockman, Greg and McLeavey, Christine and Sutskever, Ilya},
  booktitle={International conference on machine learning},
  pages={28492--28518},
  year={2023},
  organization={PMLR}
}

@article{effort,
  title={Orthogonal subspace decomposition for generalizable ai-generated image detection},
  author={Yan, Zhiyuan and Wang, Jiangming and Jin, Peng and Zhang, Ke-Yue and Liu, Chengchun and Chen, Shen and Yao, Taiping and Ding, Shouhong and Wu, Baoyuan and Yuan, Li},
  journal={arXiv preprint arXiv:2411.15633},
  year={2024}
}

@inproceedings{LSDA,
  title={Transcending forgery specificity with latent space augmentation for generalizable deepfake detection},
  author={Yan, Zhiyuan and Luo, Yuhao and Lyu, Siwei and Liu, Qingshan and Wu, Baoyuan},
  booktitle={Proceedings of the IEEE/CVF Conference on Computer Vision and Pattern Recognition},
  pages={8984--8994},
  year={2024}
}

@inproceedings{SBI,
  title={Detecting deepfakes with self-blended images},
  author={Shiohara, Kaede and Yamasaki, Toshihiko},
  booktitle={Proceedings of the IEEE/CVF conference on computer vision and pattern recognition},
  pages={18720--18729},
  year={2022}
}

@inproceedings{sladd,
  title={Self-supervised learning of adversarial example: Towards good generalizations for deepfake detection},
  author={Chen, Liang and Zhang, Yong and Song, Yibing and Liu, Lingqiao and Wang, Jue},
  booktitle={Proceedings of the IEEE/CVF conference on computer vision and pattern recognition},
  pages={18710--18719},
  year={2022}
}

@inproceedings{squeeze,
  title={Squeeze-and-excitation networks},
  author={Hu, Jie and Shen, Li and Sun, Gang},
  booktitle={Proceedings of the IEEE conference on computer vision and pattern recognition},
  pages={7132--7141},
  year={2018}
}

@inproceedings{wang2026beyond,
  title={Beyond [CLS] Token: Query-Driven Token-Level Forgery Purification for Generalizable Deepfake Detection},
  author={Wang, Changshuo and Wang, Jiangming and Zhang, Ke-Yue and Yao, Taiping and Ding, Shouhong and Wang, Shunli and Yi, Ran and Ma, Lizhuang},
  booktitle={Proceedings of the IEEE/CVF Conference on Computer Vision and Pattern Recognition},
  pages={42922--42931},
  year={2026}
}

@article{iwax,
  title={iWAX: interpretable Wav2vec-AASIST-XGBoost framework for voice spoofing detection},
  author={Lee, Seungeun and Choi, Sunmook and Kang, Taein and Chung, Sanghyeok and Han, Soyul and Seo, Jaejin and Park, Seoyoung and Kim, Eujin and Oh, Seungsang and Kwak, Il-Youp},
  journal={Scientific reports},
  volume={15},
  number={1},
  pages={40491},
  year={2025},
  publisher={Nature Publishing Group UK London}
}

@inproceedings{wpt,
  title={Detect all-type deepfake audio: Wavelet prompt tuning for enhanced auditory perception},
  author={Xie, Yuankun and Fu, Ruibo and Wang, Xiaopeng and Wang, Zhiyong and Cao, Songjun and Ma, Long and Cheng, Haonan and Ye, Long},
  booktitle={Proceedings of the AAAI Conference on Artificial Intelligence},
  volume={40},
  number={42},
  pages={35922--35930},
  year={2026}
}

@article{wav2vec+aasist,
  title={Automatic speaker verification spoofing and deepfake detection using wav2vec 2.0 and data augmentation},
  author={Tak, Hemlata and Todisco, Massimiliano and Wang, Xin and Jung, Jee-weon and Yamagishi, Junichi and Evans, Nicholas},
  journal={arXiv preprint arXiv:2202.12233},
  year={2022}
}

@article{xlsr,
  title={Unsupervised cross-lingual representation learning for speech recognition},
  author={Conneau, Alexis and Baevski, Alexei and Collobert, Ronan and Mohamed, Abdelrahman and Auli, Michael},
  journal={arXiv preprint arXiv:2006.13979},
  year={2020}
}

@article{wav2vec2.0,
  title={wav2vec 2.0: A framework for self-supervised learning of speech representations},
  author={Baevski, Alexei and Zhou, Yuhao and Mohamed, Abdelrahman and Auli, Michael},
  journal={Advances in neural information processing systems},
  volume={33},
  pages={12449--12460},
  year={2020}
}

@article{wu2017asvspoof,
  title={ASVspoof: The automatic speaker verification spoofing and countermeasures challenge},
  author={Wu, Zhizheng and Yamagishi, Junichi and Kinnunen, Tomi and Hanil{\c{c}}i, Cemal and Sahidullah, Mohammed and Sizov, Aleksandr and Evans, Nicholas and Todisco, Massimiliano and Delgado, Hector},
  journal={IEEE Journal of Selected Topics in Signal Processing},
  volume={11},
  number={4},
  pages={588--604},
  year={2017},
  publisher={IEEE}
}

@inproceedings{simlbr,
  title={SimLBR: Learning to Detect Fake Images by Learning to Detect Real Images},
  author={Dhakal, Aayush and Khanal, Subash and Sastry, Srikumar and Arndt, Jacob and Dias, Philipe and Lunga, Dalton and Jacobs, Nathan},
  booktitle={Proceedings of the IEEE/CVF Conference on Computer Vision and Pattern Recognition},
  pages={35472--35482},
  year={2026}
}

@article{lora,
  title={Lora: Low-rank adaptation of large language models.},
  author={Hu, Edward J and Shen, Yelong and Wallis, Phillip and Allen-Zhu, Zeyuan and Li, Yuanzhi and Wang, Shean and Wang, Liang and Chen, Weizhu and others},
  journal={Iclr},
  volume={1},
  number={2},
  pages={3},
  year={2022}
}

@article{fovb,
  title={Towards Generalizable deepfake detection via forgery-aware audio-visual adaptation: A variational bayesian approach},
  author={Nie, Fan and Ni, Jiangqun and Zhang, Jian and Zhang, Bin and Zhang, Weizhe and Li, Bin},
  journal={IEEE Transactions on Information Forensics and Security},
  year={2026},
  publisher={IEEE}
}

@article{mcl,
  title={Mcl: multimodal contrastive learning for deepfake detection},
  author={Liu, Xiaolong and Yu, Yang and Li, Xiaolong and Zhao, Yao},
  journal={IEEE Transactions on Circuits and Systems for Video Technology},
  volume={34},
  number={4},
  pages={2803--2813},
  year={2023},
  publisher={IEEE}
}

\end{document}